\documentclass{article}

\PassOptionsToPackage{authoryear,round}{natbib}
 \usepackage[preprint]{neurips_2025}

\usepackage[utf8]{inputenc} % allow utf-8 input
\usepackage[T1]{fontenc}    % use 8-bit T1 fonts
\usepackage{hyperref}       % hyperlinks
\usepackage{url}            % simple URL typesetting
\usepackage{booktabs}       % professional-quality tables
\usepackage{amsfonts}       % blackboard math symbols
\usepackage{nicefrac}       % compact symbols for 1/2, etc.
\usepackage{microtype}      % microtypography
\usepackage{xcolor}         % colors
\usepackage{graphicx}
\usepackage{svg}
\usepackage{colortbl}
\usepackage{enumitem}
\usepackage{wrapfig}
\usepackage{caption}

\hypersetup{
    colorlinks=true,   % <-- 这是关键！设置为 true, 文本变色, 边框消失
    citecolor=neuripsaccentblue,    % 与标题颜色保持一致
    linkcolor=neuripsaccentblue,    % 与标题颜色保持一致
    urlcolor=neuripsaccentblue      % 与标题颜色保持一致
}

\title{OpenSeeker-v2: Pushing the Limits of Search Agents with Informative and High-Difficulty Trajectories}

\author{%
  \textbf{Yuwen Du\textsuperscript{1,*}}, \textbf{Rui Ye\textsuperscript{1,*,\#,\textdagger}},  \textbf{Shuo Tang\textsuperscript{1}}, \textbf{Keduan Huang\textsuperscript{1}}, \textbf{Xinyu Zhu\textsuperscript{1}}, \textbf{Yuzhu Cai\textsuperscript{1}}, \textbf{Siheng Chen\textsuperscript{1,\textdagger}} \\
  \textsuperscript{1}Shanghai Jiao Tong University, \textsuperscript{*}Equal Core Contributions, \textsuperscript{\#}Project Lead \\
  \textsuperscript{\textdagger}Corresponding Authors: yr991129@sjtu.edu.cn, sihengc@sjtu.edu.cn
}

\begin{document}

\maketitle

\begin{abstract}
  Deep search capabilities have become an indispensable competency for frontier Large Language Model (LLM) agents, yet their development remains dominated by industrial giants.
  The typical industry recipe involves a highly resource-intensive pipeline spanning pre-training, continual pre-training (CPT), supervised fine-tuning (SFT), and reinforcement learning (RL).
  In this report, we show that when fueled with informative and high-difficulty trajectories, a simple SFT approach could be surprisingly powerful for training frontier search agents.
  By introducing three simple data synthesis modifications: scaling knowledge graph size for richer exploration, expanding the tool set size for broader functionality, and strict low-step filtering, we establish a stronger baseline.
  Trained on merely 10.6k data points, our OpenSeeker-v2 achieves state-of-the-art performance across 4 benchmarks (30B-sized agents with ReAct paradigm): \textbf{46.0\% on BrowseComp, 58.1\% on BrowseComp-ZH, 34.6\% on Humanity's Last Exam, and 78.0\% on xbench}, surpassing even Tongyi DeepResearch trained with heavy CPT+SFT+RL pipeline, which achieves 43.4\%, 46.7\%, 32.9\%, and 75.0\%, respectively.
  Notably, OpenSeeker-v2 represents the first state-of-the-art search agent within its model scale and paradigm to be developed by a purely academic team using only SFT.
  We are excited to open-source the OpenSeeker-v2 model weights and share our simple yet effective findings to make frontier search agent research more accessible to the community.

  \medskip
  \begin{flushleft}
    \begin{tabular}{@{}ll@{}}
      \includegraphics[width=1em]{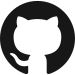} \quad \textbf{Code} & \href{https://github.com/PolarSeeker/OpenSeeker}{https://github.com/PolarSeeker/OpenSeeker} \\
      \includegraphics[width=1em]{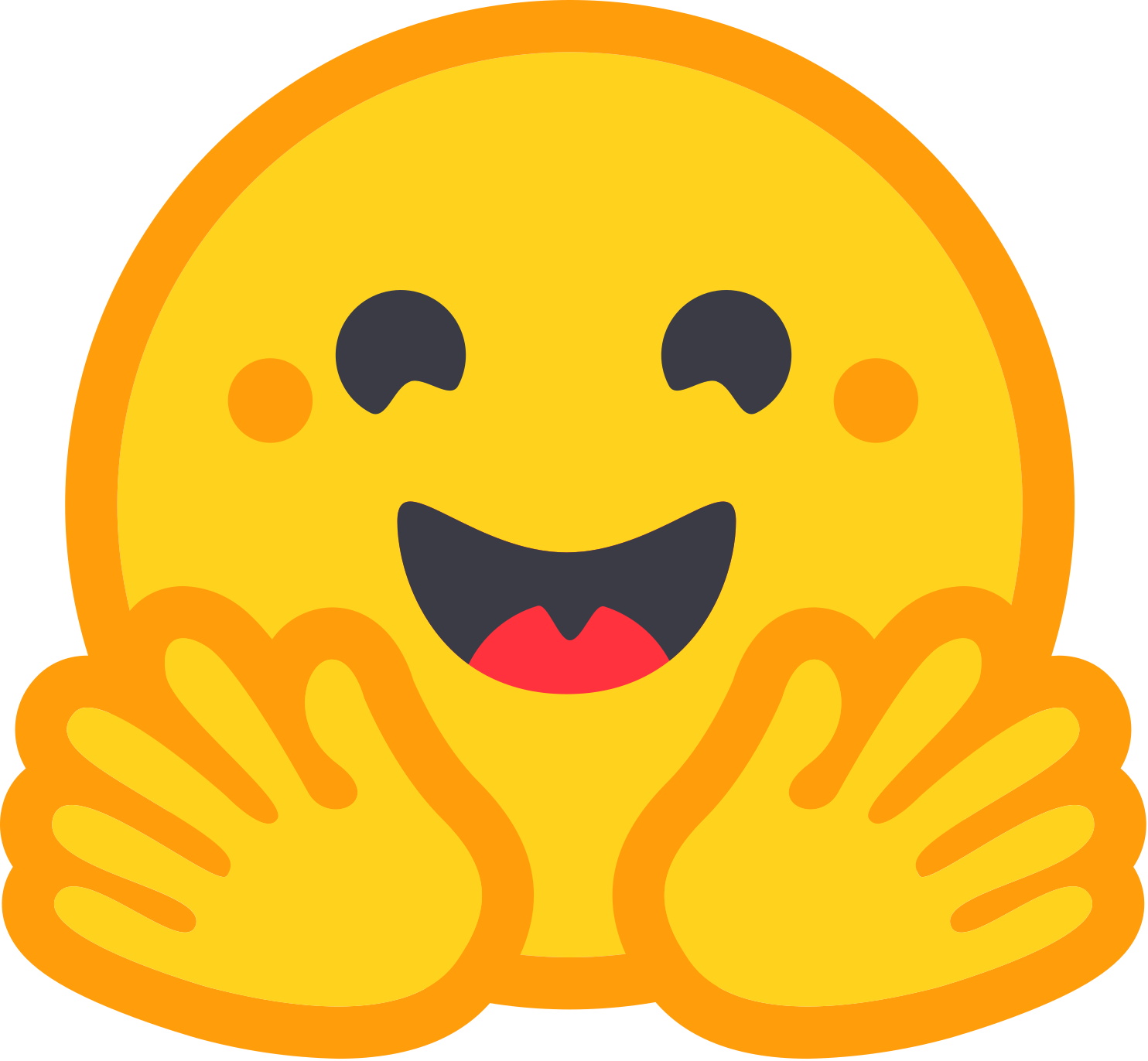} \quad \textbf{Model} & \href{https://huggingface.co/PolarSeeker/OpenSeeker-v2-30B-SFT}{https://huggingface.co/PolarSeeker/OpenSeeker-v2-30B-SFT} \\
    \end{tabular}
  \end{flushleft}
\end{abstract}

\begin{figure}[!h]
    \centering
    \vspace{-4mm}
    \includegraphics[width=1.0\linewidth]{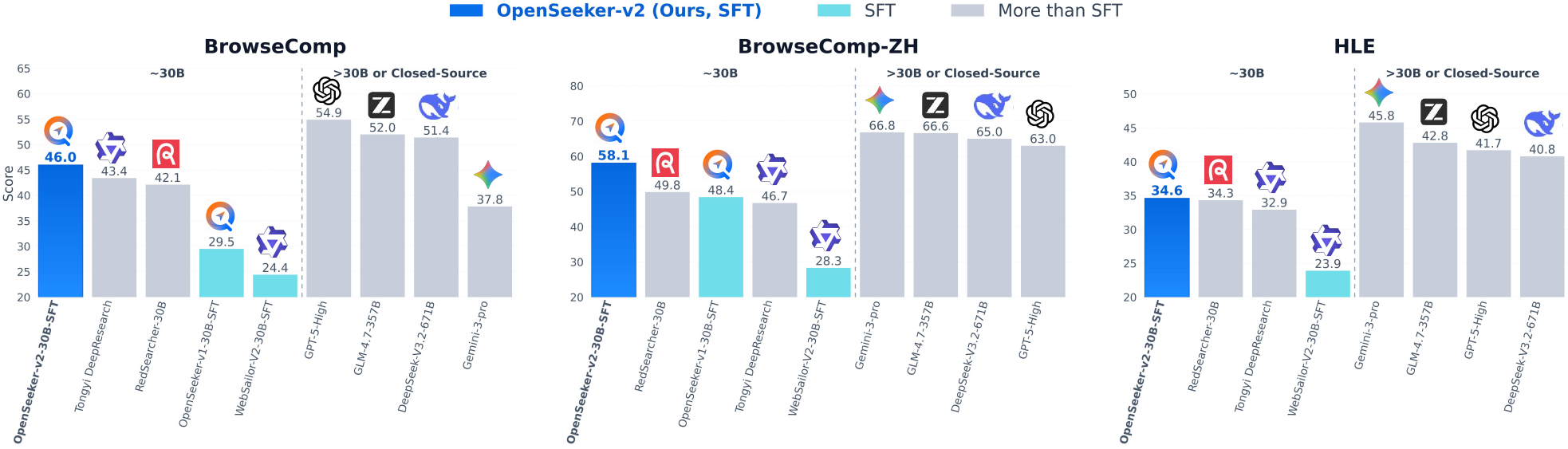}
    \vspace{-3mm}
    \caption{OpenSeeker-v2 achieves state-of-the-art performance within its model scale and paradigm on four representative benchmarks, remarkably accomplishing this via simple SFT and outperforming Tongyi DeepResearch that is trained via extensive continual pre-training, SFT, and RL.}
    \label{fig:teaser}
\end{figure}

\section{Introduction}

In the era of information explosion, deep search has emerged as a non-negotiable competency for frontier Large Language Model (LLM) agents~\citep{dr}.
However, the development of these high-performance agents has long remained a "closed-door game" played almost exclusively by well-funded corporate entities~\citep{o3,claude4}.
The typical industry recipe to achieve state-of-the-art (SOTA) performance is highly resource-intensive, typically involving Continual Pre-Training (CPT) on massive corpora~\citep{team2025tongyi,team2026mirothinker,chu2026redsearcher}, followed by Supervised Fine-Tuning (SFT)~\citep{ye2025agentfold}, and culminating in complex Reinforcement Learning (RL) stages~\citep{li2025websailorv2}.
This heavy reliance on immense compute and proprietary data pipelines has created a massive barrier, fundamentally hindering the academic and open-source communities from innovating within this domain.

We challenge this prevailing reliance on complex, multi-stage training pipelines.
Building upon our initial exploration in OpenSeeker~\citep{du2026openseeker}, we shift the focus entirely back to the quality of the training trajectories themselves and ask a crucial question:
\textit{can we push the limits of search agents and rival the performance of heavy industrial pipelines using only a straightforward SFT approach?}

In this report, we introduce OpenSeeker-v2, an upgraded search agent that proves a straightforward SFT approach could be sufficiently powerful when fueled by \textit{high-quality data of high difficulty and richness.}
Specifically, we introduce two simple yet highly effective modifications to our data synthesis pipeline:  
(1) Scaling graph size for richer exploration: We significantly expand the topological graph size during data generation.
This expansion injects a much richer and more diverse set of source information into the context, enabling the synthesis of highly complex tasks that structurally mandate deep, multi-hop exploration to solve.
(2) Expanding the tool set for broader functionality: We increase the number of available tools, allowing the agent to learn more versatile strategies and handle a wider variety of queries.
(3) Strict low-step filtering: We filter out any trajectory that can be resolved in too few tool-call steps. By intentionally dropping these simple queries, we guarantee a strict minimum difficulty floor for the training set, forcing the agent to learn sustained reasoning and information seeking over long horizons.

By applying these two strategies, we curate a highly condensed dataset of merely 10k high-difficulty trajectories.
Strikingly, training a ~30B parameter model on this small dataset via a single SFT run yields surprising results.
OpenSeeker-v2 achieves a new SOTA\footnote{While some works focus on context management~\citep{ye2025agentfold,team2026mirothinker}, our work focuses on general ReAct-based paradigm with emphasis on data quality.} across four representative agentic benchmarks: \textbf{46.0\% on BrowseComp, 58.1\% on BrowseComp-ZH, 34.6\% on Humanity's Last Exam, and 78.0\% on xbench}.
Notably, this simple SFT baseline decisively outperforms prominent industrial models such as Tongyi DeepResearch, which relies on an extensive CPT+SFT+RL pipeline and achieves 43.4\%, 46.7\%, 32.9\%, and 75.0\%, respectively.  

Ultimately, OpenSeeker-v2 represents the first state-of-the-art search agent within its model scale and paradigm (ReAct) to be developed entirely by a purely academic team using only SFT.
To democratize frontier search agent research and provide an easily reproducible baseline for the community, we are excited to fully open-source the OpenSeeker-v2 model weights.

\section{Methodology and Results}

\subsection{Methodology}

We introduce \textbf{OpenSeeker-v2}, an upgraded search-agent training framework based on supervised fine-tuning (SFT). Our central hypothesis is that, given sufficiently difficult and information-rich training data, a straightforward SFT objective is enough to induce strong long-horizon search and reasoning abilities.

\textbf{Scaling graph size for richer exploration.}
Let $\mathcal{G}=(\mathcal{V},\mathcal{E})$ denote the source graph used for task synthesis. For each seed node $v_{\mathrm{seed}}\in\mathcal{V}$, the original pipeline constructs a local subgraph $\mathcal{G}_{\mathrm{sub}}$ around $v_{\mathrm{seed}}$. In OpenSeeker-v2, we increase the expansion budget from $k$ to $K$, where $K > k$, and obtain a larger evidence subgraph:
\[
\mathcal{G}_{\mathrm{sub}}^{(K)}
=
\operatorname{Expand}(\mathcal{G}, v_{\mathrm{seed}}, K).
\]
The enlarged subgraph contains a richer set of topologically related sources, which increases the number and diversity of feasible reasoning paths. A synthetic query is then generated conditioned on this expanded context:
\[
q \sim P_{\mathrm{gen}}\left(q \mid \mathcal{G}_{\mathrm{sub}}^{(K)}\right).
\]
By scaling $K$, the generated question is more likely to require evidence aggregation over multiple nodes rather than relying on few source.

\textbf{Expanding the tool set for broader functionality.}
Given a generated question $q$, we equip the search agent with an expanded set of tools $\mathcal{A}$ larger than that used in OpenSeeker-v1~\citep{du2026openseeker} following~\cite{team2026mirothinker} and let it produce a multi-step ReAct-style trajectory:
\[
\tau =
\left(
r_1,a_1,o_1,
r_2,a_2,o_2,
\ldots,
r_T,a_T,o_T,
r_{T+1},y
\right),
\]
where each action $a_t \in \mathcal{A}$ corresponds to a tool call selected from the enlarged tool set, and $o_t$ denotes the observation returned by the invoked tool.
$r_t$ represents the reasoning trace before each action.
The trajectory consists of $T$ tool-call steps, followed by a final reasoning step $r_{T+1}$ and the answer $y$.
By expanding $\mathcal{A}$, the agent is encouraged to learn more diverse interaction patterns and leverage complementary tools, resulting in more flexible and functionally rich problem-solving behaviors.

\textbf{Strict low-step filtering.}
To remove overly simple instances, we apply a strict low-step filtering rule:
\[
\mathcal{D}_{\mathrm{v2}}
=
\left\{
(q,\tau)
\in
\mathcal{D}_{\mathrm{raw}}
\;\middle|\;
T(\tau) \geq T_{\min}
\right\}.
\]
Here, $T_{\min}$ is a predefined minimum tool-call threshold. Trajectories with $T(\tau)<T_{\min}$ are discarded because they can often be solved by direct lookup or shallow keyword matching.

Finally, OpenSeeker-v2 trains the search agent with a standard SFT objective over the filtered dataset.

The expanded graph increases contextual richness and multi-hop dependency, while low-step filtering enforces a minimum difficulty floor. Together, these two modifications produce high-quality SFT data that encourages the agent to learn sustained reasoning, robust information extraction, and long-horizon search behavior.

\subsection{Experimental Setup}

\textbf{Implementation.}
We instantiate OpenSeeker-v2 from Qwen3-30B-A3B-Thinking-2507~\citep{qwen3thinking}, which has 30B total parameters and 3B activated parameters during inference. The agent uses a 256k context window and allows up to 200 tool calls per trajectory. OpenSeeker-v2 is trained with SFT, without RL or additional hyperparameter tuning.

\textbf{Benchmarks.}
We evaluate OpenSeeker-v2 on five challenging agentic benchmarks: BrowseComp~\citep{bc_en}, BrowseComp-ZH~\citep{bc_zh}, Humanity's Last Exam (HLE)~\citep{hle}, and xbench-DeepSearch~\citep{xbench}. 
These benchmarks cover diverse deep research tasks.
We mask the hugging-face-related links when calling the web search tools to avoid potential leakage.

\textbf{Baselines.}
We compare OpenSeeker-v2 with representative systems in Table~\ref{tab:deep_research_complete}, with a primary focus on comparable-scale ReAct-based search agents. Tongyi DeepResearch~\citep{team2025tongyi} and RedSearcher~\citep{chu2026redsearcher} are strong 30B-scale search agents trained with heavier CPT+SFT+RL pipelines. They provide direct references for evaluating whether our SFT-only approach can rival more resource-intensive training recipes. For completeness, we also include closed-source proprietary models~\citep{claude4,o3,dr,Singh2025GPT5SystemCard} and large open-source models~\citep{DeepSeek2025V32,5team2025glm45agenticreasoningcoding,MiniMaxM2} as broader reference points. Baseline results are taken from their technical reports or public leaderboards.

\subsection{Main Results}
\begin{table*}[t]
    \centering
    \small
        \caption{Comparisons among our OpenSeeker and other ReAct-based search agents. `\# Samples' denotes the number of total training data samples; `Training' denotes training techniques (CPT: continual pre-training, SFT: supervised fine-tuning, RL: reinforcement learning); `Academic' denotes whether conducted by pure academic team ($\checkmark$: Yes, $\times$: No); `BC-ZH' denotes BrowseComp-ZH. Notably, with simple SFT only, OpenSeeker-v2-30B-SFT consistently outperforms models of comparable scale trained with more complex pipelines involving CPT, SFT, and RL. \textbf{OpenSeeker-v2 comprehensively outperforms pure ReAct-based models of comparable scale.}}
        \label{tab:deep_research_complete}
        \resizebox{\textwidth}{!}{
        \begin{tabular}{lccccccc}
            \toprule
            \textbf{Model Name} & \textbf{\# Samples} & \textbf{Training} & \textbf{Academic} & \textbf{BrowseComp} & \textbf{BC-ZH} & \textbf{HLE} & \textbf{xbench}\\
            \midrule
            
            % --- Category 1 ---
            \rowcolor{blue!8}\multicolumn{8}{c}{\emph{\textbf{Closed-Source Proprietary Models}}} \\
            Claude-4-Opus & ? & ? & $\times$ & 18.8 & 37.4 & - & - \\
            Claude-4.5-Sonnet & ? & ? & $\times$ & 24.1 & 42.4 & 32.0 & - \\
            Gemini-3-pro & ? & ? & $\times$ & 37.8 & 66.8 & 45.8 & - \\
            OpenAI-o3 & ? & ? & $\times$ & 49.1 & 68.7 & 20.2 & 65.0 \\
            OpenAI Deep Research & ? & ? & $\times$ & 51.5 & 42.9 & 26.6 & -  \\
            GPT-5-High & ? & ? & $\times$ & 54.9 & 63.0 & 41.7 & - \\
            \midrule

            % --- Category 2 ---
           \rowcolor{blue!8}\multicolumn{8}{c}{\emph{\textbf{Open-Source Models > 30B}}} \\
            DeepSeek-V3.1-671B & ? & ? & $\times$ & 30.0 & 49.2 & 29.8 & 71.2 \\
            DeepSeek-V3.2-671B & ? & ? & $\times$ & 51.4 & 65.0 & 40.8 & -  \\
            GLM-4.6-357B & ? & ? & $\times$ & 45.1 & 49.5 & 30.4 & - \\
            GLM-4.7-357B & ? & ? & $\times$ & 52.0 & 66.6 & 42.8 & - \\
            Minimax-M2-230B & ? & ? & $\times$ & 44.0 & 48.5 & - & - \\
            \midrule

            % --- Category 3 ---
            \rowcolor{blue!8}\multicolumn{8}{c}{\emph{\textbf{$\sim$30B Models}}} \\
            WebSailor-V2-30B-SFT & ? & SFT & $\times$ & 24.4 & 28.3 & 23.9 & 61.7 \\
            WebSailor-V2-30B-RL & ? & SFT + RL & $\times$ & 35.3 & 44.1 & 30.6 & 73.7 \\
            WebLeaper-30B-SFT & 15 k & SFT & $\times$ &27.7 & - & - & 66.0 \\
            WebLeaper-30B-RL & ? & RL & $\times$ &38.8 & - & - & 72.0 \\
            Tongyi DeepResearch & ? & CPT + SFT + RL & $\times$ & 43.4 & 46.7 & 32.9 & 75.0 \\
            RedSearcher-30B & ? & CPT + SFT + RL & $\times$ &42.1  & 49.8 & 34.3 & - \\
            OpenSeeker-v1-30B-SFT & 11.7 k & SFT & \checkmark & 29.5 & 48.4 & - & 74.0 \\
            \cmidrule(lr){1-8}
            
            \textbf{OpenSeeker-v2-30B-SFT} & 10.6 k & SFT & \checkmark & \textbf{46.0} & \textbf{58.1} & \textbf{34.6} & \textbf{78.0} \\
            \bottomrule
        \end{tabular}
        }
\end{table*}

\textbf{Surpassing comparable-scale agents trained with heavier pipelines.}
The central question behind OpenSeeker-v2 is whether simple SFT can push the limits of search agents and rival heavier industrial pipelines. As shown in Table~\ref{tab:deep_research_complete}, OpenSeeker-v2-30B-SFT achieves the strongest overall performance among $\sim$30B ReAct-based search agents while using SFT only. 
OpenSeeker-v2 achieves \textbf{46.0\% on BrowseComp, 58.1\% on BrowseComp-ZH, 34.6\% on Humanity's Last Exam, and 78.0\% on xbench}.
(1) Notably, with simple SFT, OpenSeeker-v2 outperforms Tongyi DeepResearch developed by Alibaba Tongyi Lab~\citep{team2025tongyi} and RedSearcher developed by RedNote, which are trained by the extensive CPT+SFT+RL pipeline
Specifically, on the challenging benchmarks BrowseComp and HLE, OpenSeeker-v2 outperforms these two by at least 2.6\% and 0.3\%, respectively; while on the BrowseComp-ZH and xbench, OpenSeeker-v2 significantly outperforms Tongyi DeepResearch by 11.4\% and 3\%, respectively.
(2) Comparing with larger models, OpenSeeker-v2 also outperforms DeepSeek-V3.1-671B, GLM-4.6-357B, Minimax-M2-230B, Claude-4.5-Sonnet, indicating its strong capability.
These results demonstrate that a straightforward SFT approach can be sufficiently powerful when fueled by high-quality data of high difficulty and richness, suggesting that data quality could be a critical path towards training intelligent long-horizon search agents.

\textbf{Demonstrating the scaling potential of OpenSeeker.}
OpenSeeker-v2 substantially improves upon OpenSeeker-v1~\citep{du2026openseeker} under the same model scale and SFT-only training recipe, highlighting the development potential of the OpenSeeker framework through higher-quality data construction.
OpenSeeker-v2 raises BrowseComp from 29.5 to 46.0, BrowseComp-ZH from 48.4 to 58.1, xbench from 74.0 to 78.0. These gains suggest that OpenSeeker has not yet saturated under the current SFT setting.
More importantly, they show that increasing the difficulty and richness of synthesized QA tasks and enhancing the overall quality of synthesized trajectories can lead to substantial capability gains, indicating that scalable high-quality data synthesis is a promising path for further advancing search agents.

\setlength{\intextsep}{0pt}
\setlength{\columnsep}{0.6em}

\begin{wrapfigure}[11]{r}{0.48\textwidth}
    \centering
    \includegraphics[width=\linewidth]{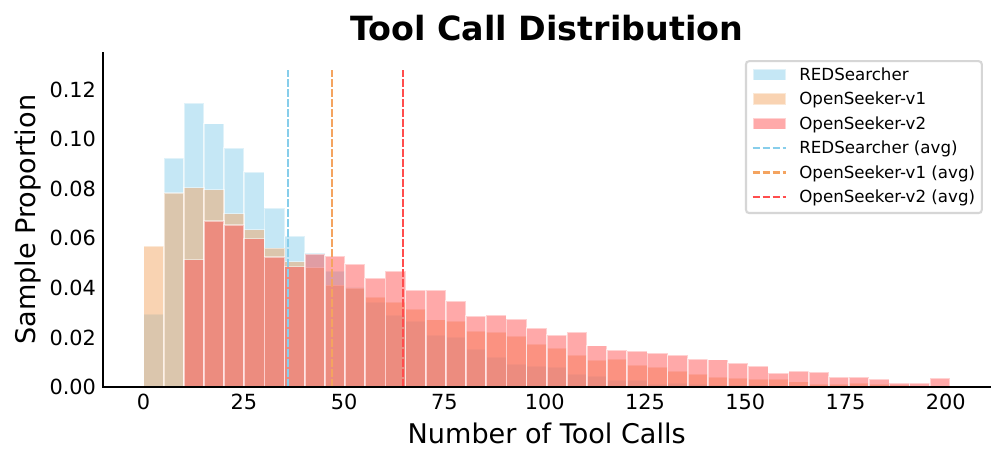}
    \vspace{-1.5em}
    \captionsetup{width=\linewidth}
    \caption{
        Comparison of average tool
call counts across search-agent training data.
    }
    \label{fig:trajectory_length_comparison}
\end{wrapfigure}

\textbf{OpenSeeker-v2 demonstrates higher data difficulty than prior counterparts.}
OpenSeeker-v2 is built upon substantially longer search trajectories, with an average of
64.67 steps per trajectory, compared with 46.97 for OpenSeeker-v1
and 36.01 for RedSearcher. This suggests that the OpenSeeker-v2 training data
requires more complex multi-step reasoning and longer-horizon information seeking.
We hypothesize that such long and difficult synthetic trajectories are crucial for
enabling the model to acquire stronger long-horizon retrieval and search capabilities,
which further explains the superior performance of OpenSeeker-v2 on challenging
deep-research benchmarks.

\section{Conclusion}

In this report, we share that when fueled by high-quality data of high-difficulty and richness, a search agent trained with simple SFT could rival the performance of agents trained with extensive resources.
Specifically, we share three simple yet effective modifications on the data collection pipeline: scaling graph size, expanding tool set, and low-step filtering, and train our final search agent: OpenSeeker-v2.
Though trained with only 10.6k samples, OpenSeeker-v2 achieves a new SOTA across four representative benchmarks: 46.0\% on BrowseComp, 58.1\% on BrowseComp-ZH, 34.6\% on Humanity’s Last Exam, and 78.0\% on xbench, significantly outperforms Tongyi DeepResearch and RedSearcher that are extensively trained via CPT, SFT, and RL.
Our report highlight the critical role of data quality, suggesting that carefully designed data alone can unlock substantial performance gains.

\textbf{What's next.}
Our internal observations suggest strong scaling potential of high-quality synthesized data. 
Moving forward, we will continue to push in this direction by scaling up data quantity, quality, and diversity, with the goal of further pushing the limits of search agents.

\medskip
{
\bibliographystyle{plainnat}
\bibliography{ref}
}

\clearpage
\appendix

\end{document}